# Decision-Theoretic Foundations for Causal Reasoning

**David Heckerman**    HECKERMA@MICROSOFT.COM
*Microsoft Research, One Microsoft Way*
*Redmond, WA 98052-6399 USA*

**Ross Shachter**    SHACHTER@CAMIS.STANFORD.EDU
*Stanford University*
*Stanford, CA 94305-4025 USA*

## Abstract

We present a definition of cause and effect in terms of decision-theoretic primitives and thereby provide a principled foundation for causal reasoning. Our definition departs from the traditional view of causation in that causal assertions may vary with the set of decisions available. We argue that this approach provides added clarity to the notion of cause. Also in this paper, we examine the encoding of causal relationships in directed acyclic graphs. We describe a special class of influence diagrams, those in canonical form, and show its relationship to Pearl's representation of cause and effect. Finally, we show how canonical form facilitates counterfactual reasoning.

## 1. Introduction

Knowledge of cause and effect is crucial for modeling the affects of actions. For example, if we observe a statistical correlation between smoking and lung cancer, we can not conclude from this observation alone that our chances of getting lung cancer will change if we stop smoking. If, however, we also believe that smoking is a cause for lung cancer, then we can conclude that our choice whether to continue or quit smoking will affect whether we get lung cancer.

Work by artificial intelligence researchers, statisticians, and philosophers have emphasized the importance of identifying causal relationships for purposes of modeling the effects of actions. For example, Simon (1977), Robins (1986), Spirtes et al. (1993), and Pearl (1993, 1995) have developed graphical models of cause and effect, and have demonstrated how these models are important for reasoning about the effects of actions. In addition, Robins (1986), Rubin (1978), Pearl and Verma (1991), and Spirtes et al. (1993) have developed approaches that embrace causality for learning the effects of actions from data.

One useful framework for causal reasoning is that of Pearl (1993, 1995)—herein Pearl. Using his framework, we construct a *causal graph* $G$. The nodes in $G$ correspond to a set of variables $U$ that we wish to model. Each variable has a set of mutually exclusive and collectively exhaustive values or *instances*. The arcs in $G$ represent (informal) assertions of cause—in particular, the parents of $x \in U$ are direct causes of $x$. Pearl gives these informal assertions of cause an operational meaning by introducing a special class of actions on the variables $U$ and then describing the affects of these actions using the structure of the causal graph. Specifically, he posits that, for every variable $x \in U$, there exists another variable $\hat{x}$, which we call an *atomic intervention on* $x$. The variable $\hat{x}$ has an instance **set(x)** for every instance **x** of $x$, and an instance **idle**. The instance **set(x)** corresponds to an action that forces $x$ to take on instance $\mathbf{x}$ and indirectly affects other variables through the change in $x$. The instance **idle** corresponds to the action of doing nothing.[1] Pearl then asserts that the effects of atomic interventions on the variables in $U$ are determined by the *structural equations*

$$x = f_x(Pa^G(x), \hat{x}, \epsilon_x)$$

for all $x \in U$, where (1) $Pa^G(x)$ are the parents of $x$ in $G$—that is, the direct causes of $x$, (2) the variables $\epsilon_x$ are *exogenous* and mutually independent *random disturbances*, and (3) the function $f_x$ has the property that $x = \mathbf{x}$ when $\hat{x} =$**set**$(\mathbf{x})$ regardless of the values of $Pa^G(x)$ and $\epsilon_x$. Following Pearl, we call this framework for defining cause a *structural-equation model*.

Another useful framework for causal reasoning, closely related to Pearl's, is that of Spirtes et al. (1993)—herein SGS.

Despite these and other important advances in reasoning about cause and effect, foundations for such approaches are lacking. In any framework for causal reasoning, it is important to consider what concepts are *primitive*—that is, assumed to be self evident and used to define other concepts. As much as is possible, these primitives should have simple and universal meanings so that claims of causation can be empirically tested and causal inferences can be trusted. Unfortunately, the primitives used by Pearl, SGS, and other researchers are not ideal in this respect.

For example, SGS take cause itself to be a primitive. Given the controversies in statistics and other disciplines concerning the meaning of cause, we believe that a better primitive can be found. Pearl takes random disturbance, exogenous variable, and atomic intervention as primitives. One problem with this approach is that we need an understanding of cause and effect to identify an intervention as atomic. To illustrate the problem, suppose we wish to model the causal relationship between the binary variables $w$ and $h$ representing whether or not a person considers himself to be wealthy and happy, respectively. Further, suppose we can give this person a large sum of stolen money along with the knowledge that this money is stolen. Now we ask the question: Is this action an instance of an atomic intervention for $w$? If this person does not care about how he becomes wealthy, then the answer is "yes." If this person is more typical, however, then the answer is "no," because this action would affect both $w$ and $h$ directly. Thus, we must first determine whether or not the action is a direct cause of $h$ to determine whether or not this action is an instance of an atomic intervention.

In this paper, we provide a principled foundation for causal reasoning. In particular, we explicate a set of primitives from decision theory, and use these primitives to define the concepts of cause and atomic intervention as well as those of random disturbance and exogenous variable. These primitives are simple to understand and used uniformly across many disciplines.

The basic idea behind our definition of cause is as follows. Following the paradigm of decision theory, we focus on a person—the decision maker—who has one or more decisions to make. For each variable that we wish to model in considering these decisions, we distinguish the variable as being either a decision variable or chance variable. A *decision variable* is a variable whose instances correspond to possible actions among which the person can choose. A *chance variable* is any other variable. This framework is similar to Pearl's, where chance variables correspond to the variables $U$ and decision variables correspond to interventions. The differences are that (1) we do not require there to be a decision variable for every chance variable, and (2) decision variables need not be atomic interventions.

---

1. We adopt a variant of Pearl's notation for the instances of an atomic intervention. In addition, whereas Pearl calls the action **set**($\mathbf{x}$) an atomic intervention, we find it convenient to use this phrase to refer to the entire variable $\hat{x}$.

Now, for simplicity, suppose that we have a model consisting of only one decision variable $d$ and a set of chance variables $U$. Imagine that we choose one of the instances of $d$ and subsequently observe $x \in U$. If we believe that $x$ can be different for different choices, then, by our definition, we say that $d$ is a cause for $x$. For example, suppose decision variable $s$ represents the decision of whether or not to continue smoking and chance variable $l$ represents whether or not we get lung cancer. If we believe that we will get lung cancer if we continue smoking and that we may not get lung cancer if we quit, then we can say that $s$ is a cause of $l$.

Our definition is related to the notion of a *counterfactual*: a hypothetical statement or question that can not be verified or answered through observation (Lewis, 1973; Holland, 1986). In our smoking example, we ask the question "Will deciding differently possibly change our health outcome?" This question can not be answered by any observation, because we must either quit or continue to smoke; we can not do both. Using counterfactuals, Rubin (1978) defines a notion of *causal effect* that is closely related to our definition of cause.

The problem with most definitions of cause based on intervention is that they do not allow chance variables to be the causes of other chance variables. Consider the variables $g$ and $c$ representing a person's gender at birth and whether or not that person gets breast cancer, respectively. Although $g$ is a chance variable (we cannot choose our gender), we often hear people say in natural discourse that $g$ causes $c$. In general, we would like to accommodate such assertions.

The definition of cause that we present does indeed permit chance variables to be causes. There is, however, one catch. Namely, when we assert that a set of chance variables $X$ is a cause of chance variable $y$, we must also specify the decision or decisions that bring about the possible changes in $X$ and $y$. In our breast-cancer problem, we can assert that $g$ causes $c$, but we must explicate a decision that possibly leads to a change in gender and breast cancer. For example, we can say that $g$ causes $c$ with respect to decision variable $d$, where $d$ represents the decision of whether or not to perform genetic surgery at conception.

By including a decision context in assertions of cause, our definition departs from the traditional view of causation. Nonetheless, this departure makes causal assertions more precise. For example, consider another decision that will likely lead to a change in gender: a decision $o$ of whether or not to have a sex-change operation at birth. In this case, it may be reasonable to assert that $g$ is not a cause of $c$ with respect to $o$. Thus, causal relationships among chance variables may depend on the decisions available for intervention; and our definition accommodates this dependence.

Our paper is organized into four parts. In part 1 (Sections 2 and 3), we develop our definition of cause, using the decision-theoretic primitives of Savage (1954). In Section 2, we introduce a simpler relation than cause, which we call limited unresponsiveness. In Section 3, we define cause in terms of limited unresponsiveness.

In part 2 (Sections 4 through 7), we address the graphical representation of cause. In Section 4, we review a directed-acyclic-graph (DAG) representation, known as an *influence diagram*, which has been used for two decades by decision analysts to model the effects of decisions (Howard and Matheson, 1981). We demonstrate the inadequacies of the influence diagram as a representation of cause. In the following three sections, we develop a special condition on the influence diagram, known as *canonical form*, that improves the representation of cause.

In part 3 (Section 8), we use our definitions of cause, atomic intervention, and mapping variable, along with canonical form to build a correspondence with (and thus a foundation for) Pearl's causal framework.

In part 4 (Section 9), we demonstrate an important use of canonical form. Namely, we show how to use influence diagrams in canonical form to do general counterfactual reasoning.

We present our framework in the traditional decision-analytic paradigm of a "one shot" decision. In particular, we do not consider experimental studies, where variables are measured repeatedly. Nonetheless, one can easily extend our framework to such situations by introducing the assumption of exchangeability (de Finetti, 1937). Bayesian methods for learning models of cause that are based on this approach are discussed in Angrist et al. (1995) and Heckerman (1995).

## 2. Unresponsiveness

In this section, we introduce the notion of limited unresponsiveness, a fundamental relation that we use to define cause. We define limited unresponsiveness using primitives from decision theory as explicated (for example) by Savage (1954).

We begin with a description of the primitives *act*, *consequence*, and *possible state of the world*. Savage describes and illustrates these concepts as follows:

> To say that a decision is to be made is to say that one or more acts is to be chosen, or decided on. In deciding on an act, account must be taken of the possible states of the world, and also of the consequences implicit in each act for each possible state of the world. A *consequence* is anything that may happen to the person.
>
> Consider an example. Your wife has just broken five good eggs into a bowl when you come in and volunteer to finish making the omelet. A sixth egg, which for some reason must either be used for the omelet or wasted altogether, lies unbroken beside the bowl. You must decide what to do with this unbroken egg. Perhaps it is not too great an oversimplification to say that you must decide among three acts only, namely, to break it into the bowl containing the other five, to break it into a saucer for inspection, or to throw it away without inspection. Depending on the state of the egg, each of these three acts will have some consequence of concern to you, say that indicated by Table 1.

For purposes of our discussion, there are two points to emphasize from Savage's exposition. First, it is important to distinguish between that which we can choose—namely, acts—and that which we can not choose—namely, consequences. Second, once we choose an act, the consequence that occurs is logically determined by the state of the world. That is, the consequence is a deterministic function of the act and the state of the world. Of course, the consequence can be (and usually is) uncertain, and this uncertainty is captured by uncertainty in the state of the world. The concepts of act, consequence, and state of the world, together with the deterministic mapping from act and state of the world to consequence are our only primitives.[2]

| state of the world | act | | |
|---|---|---|---|
| | break into bowl | break into saucer | throw away |
| good egg | six-egg omelet | six-egg omelet and a saucer to wash | five-egg omelet and one good egg destroyed |
| bad egg | no omelet and five good eggs destroyed | five-egg omelet and a saucer to wash | five-egg omelet |

Table 1: An example illustrating acts, possible states of the world, and consequences. (Taken from Savage [1954].)

In the omelet story, the possible states of the world readily come to mind given the description of the problem. Furthermore, we can observe the state of the world (i.e., the condition of the egg). In many if not most situations, however, the state of the world is unobservable. That is, the assertion "the state of the world is **x**" is a counterfactual. In these situations, we can bring the possible states to mind by thinking about the acts and consequences. For example, suppose we have a decision to continue smoking or quit, and we model the consequences of getting cancer or not. These acts and consequences bring to mind four possible states of the world, as shown in Table 2. These possible states have no familiar names; and we simply label them with numbers. The actual state of the world is not observable, because, if we decide to quit, then we won't know for sure what would have happened had we continued, and vice versa.

The acts and consequences in this problem may actually bring to mind more than four—even an infinite number—of states of the world. For example, the state of the world may correspond to degree of susceptibility of lung tissue to tar as measured by a biochemical assay. Nonetheless, given the discrete acts and consequences that we have chosen to model in the problem, the four states in Table 2 are sufficiently detailed. Savage recognizes this issue of detail in his definition of state of the world: "a description of the world, leaving no relevant aspect undescribed." In general, if we have a decision problem with $c$ consequences and $a$ acts, then at most $c^a$ possible states of the world need be distinguished.[3]

The idea that the state of the world may not be observable can be traced to Neyman (1923), who derived statistical methods for estimating the differences in yields of different crops planted on the same plot of land, in circumstances where only one crop was actually planted on a plot. Rubin (1978) and Howard (1990) have formalized this idea.

---

2. Savage (1954) *defines* an act to be "a function attaching a consequence to each state of the world." In contrast, we take act to be a primitive, as do many decision analysts (e.g., Howard, 1990).

3. When acts and consequences are continuous, the specification of $S$ is more complicated. In this paper, we address only situations where acts and consequences are discrete.

| state of the | act | |
| world | continue | quit |
|---|---|---|
| 1 | cancer | no cancer |
| 2 | no cancer | no cancer |
| 3 | cancer | cancer |
| 4 | no cancer | cancer |

Table 2: The four possible states of the world for a decision to continue or quit smoking.

In practice, it is often cumbersome if not impossible to reason about a monolithic set of acts, possible states of the world, or consequences. Therefore, we typically describe each of these items in terms of a set of variables that take on two or more values or *instances*. We call a variable describing a set of consequences a *chance variable*. For example, in the omelet story, we can describe the consequences in terms of three chance variables: (1) *number of eggs in the omelet?*[4] (*o*) having instances **zero**, **five**, and **six**, (2) *number of good eggs destroyed?* (*g*) having instances **zero**, **one**, and **five**, and (3) *saucer to wash?* (*s*) having instances **no** and **yes**. That is, every consequence corresponds to an assignment of an instance to each chance variable.

We call a variable describing a set of acts a *decision variable* (or *decision*, for short). For example, suppose we have a set of possible acts about how we are going to dress for work. In this case, we can describe the acts in terms of the decision variables *shirt* (**plain** or **striped**), *pants* (**jeans** or **corduroy**), and *shoes* (**tennis shoes** or **loafers**). In this example and in general, every act corresponds to a choice of an instance for each decision variable.

The description of possible states of the world in terms of component variables is a bit more complicated, and is not needed for our explication of unresponsiveness and limited unresponsiveness. We defer discussion of this issue to Section 6.

As a matter of notation, we use $D$ to denote the set of decisions that describe the acts for a decision problem, and lower-case letters (e.g., $d, e, f$) to denote individual decisions in the set $D$. Also, we use $U$ to denote the set of chance variables that describe the consequences, and lower-case letters (e.g., $x, y, z$) to denote individual chance variables in $U$. In addition, we use the variable $S$ to denote the state of the world (the instances of $S$ correspond to the possible states of the world).[5] Thus, any given decision problem—or *domain*, as we sometimes call it—is described by the variables $U$, $D$, and $S$.[6]

With this introduction, we can discuss the concept of limited unresponsiveness. To illustrate this concept, consider the following decision problem adapted from Angrist et al. (1995). Suppose we are a physician who has to decide whether to recommend for or against a particular treatment. Given our recommendation, our patient may or may not actually accept the treatment, and may or may not be cured as a result. Here, we use a single decision variable *recommendation* ($r$) to represent our acts (i.e., $D = \{r\}$), and two chance variables *taken?* ($t$) and *cured?* ($c$) to represent whether or not the patient actually accepts the treatment and whether or not the patient is cured, respectively (i.e., $U = \{t, c\}$).

---

4. To emphasize the distinction between chance and decision variables, we put a question mark at the end of the names of chance variables.
5. We use an uppercase "$S$" to denote this single variable, because later we decompose $S$ into a set of variables.
6. Sometimes, for simplicity, we leave $S$ implicit in the specification of a decision problem.

|  | r (*recommendation*) | | | |
| S (state of the world) | take | | don't take | |
|  | t (*taken?*) | c (*cured?*) | t (*taken?*) | c (*cured?*) |
| 1: complier, helped | yes | yes | no | no |
| 2: complier, hurt | yes | no | no | yes |
| 3: complier, always cured | yes | yes | no | yes |
| 4: complier, never cured | yes | no | no | no |
| 5: defier, helped | no | no | yes | yes |
| 6: defier, hurt | no | yes | yes | no |
| 7: defier, always cured | no | yes | yes | yes |
| 8: defier, never cured | no | no | yes | no |
| 9: always taker, cured | yes | yes | yes | yes |
| 10: always taker, not cured | yes | no | yes | no |
| 11: never taker, not cured | no | no | no | no |
| 12: never taker, cured | no | yes | no | yes |
| 13: (impossible) | yes | yes | yes | no |
| 14: (impossible) | yes | no | yes | yes |
| 15: (impossible) | no | no | no | yes |
| 16: (impossible) | no | yes | no | no |

Table 3: A decision problem about recommending a medical treatment.

The possible states of the world for this problem are shown in Table 3. For example, consider the first row in the table. Here, the patient will accept the treatment if and only if we recommend it, and will be cured if and only if he takes the treatment. We describe this state by saying that the patient is a complier and is helped by the treatment. We discuss the description of these states in more detail in Section 6.

As is indicated in the table, suppose that we believe the last four states of the world are impossible (i.e., have a probability of zero). These last four states share the property that $t$ takes on the same instance for both acts, whereas $c$ does not. Thus, this decision problem satisfies the following property: in all of the states of the world that are possible, if $t$ is the same for the two acts, then $c$ is also the same. We say that $c$ is unresponsive to $r$ in states limited by $t$.

In general, suppose we have a decision problem described by variables $U$, $D$, and $S$. Let $X$ be a subset of $U$, and $Y$ be a subset of $U \cup D$. We say that $X$ is unresponsive to $D$ in states limited by $Y$ if we believe that, for all possible states of the world, if $Y$ assumes the same instance for any two acts then $X$ must also assume the same instance for those acts. We describe the notion of limited unresponsiveness in earlier work in terms of a *conditional fixed set* (Heckerman and Shachter, 1994). Angrist et al. (1995) discuss an instance of limited unresponsiveness, which they call the *exclusion restriction*.

To be more formal, let $X[\mathbf{S}, \mathbf{D}]$ be the instance that $X$ assumes (with certainty) given the state of the world $\mathbf{S}$ and the act $\mathbf{D}$. For example, in the omelet story, if $\mathbf{S}$ is the state of the world where the egg is good, and $\mathbf{D}$ is the act **throw away**, then $o[\mathbf{S}, \mathbf{D}]$ (the number of eggs in the omelet) assumes the instance **five**. Then, we have the following definition.

**Definition 1 (Limited (Un)responsiveness)** *Given a decision problem described by chance variables $U$, decision variables $D$, and state of the world $S$, and variable sets $X \subseteq U$ and $Y \subseteq D \cup U$, $X$ is said to be unresponsive to $D$ in states limited by $Y$, denoted $X \not\hookleftarrow_Y D$, if we believe that*

$$\forall\ \mathbf{S} \in S,\ \mathbf{D_1} \in D,\ \mathbf{D_2} \in D:\quad Y[\mathbf{S}, \mathbf{D_1}] = Y[\mathbf{S}, \mathbf{D_2}] \Longrightarrow X[\mathbf{S}, \mathbf{D_1}] = X[\mathbf{S}, \mathbf{D_2}]$$

$X$ is said to be responsive to $D$ in states limited by $Y$, denoted $X \hookleftarrow_Y D$, if it is not the case that $X$ is unresponsive to $D$ in states limited by $Y$—that is, if we believe that

$$\exists\ \mathbf{S} \in S,\ \mathbf{D_1} \in D,\ \mathbf{D_2} \in D\ \text{s.t.}\ Y[\mathbf{S}, \mathbf{D_1}] = Y[\mathbf{S}, \mathbf{D_2}]\ \text{and}\ X[\mathbf{S}, \mathbf{D_1}] \neq X[\mathbf{S}, \mathbf{D_2}]$$

When $X$ is (un)responsive to $D$ in states limited by $Y = \emptyset$, we simply say that $X$ *is (un)responsive to $D$*. The notion of unresponsiveness is significantly simpler than that of limited unresponsiveness. That is, when $Y = \emptyset$, the equalities on the left-hand-side of the implications in Definition 1 are trivially satisfied. Thus, $X$ is unresponsive to $D$ if we believe that, in each possible state of the world, $X$ assumes the same instance for all acts; and $X$ is responsive to $D$ if there is some possible state of the world where $X$ differs for two different acts.

As examples of responsive variables, consider the omelet story. Let $\mathbf{S}$ denote the state where the egg is good, and $\mathbf{D_1}$ and $\mathbf{D_2}$ denote the acts **break into bowl** and **throw away**, respectively. Then, for the variable $o$ (*number of eggs in omelet?*), we have $o[\mathbf{S}, \mathbf{D_1}] =$**six** and $o[\mathbf{S}, \mathbf{D_2}] =$**five**. Consequently, $o$ is responsive to $D$.[7] In a similar manner, we can conclude that $g$ (*number of good eggs destroyed?*), and $s$ (*saucer to wash?*) are each responsive to $D$.

Note that if a chance variable $x$ is responsive to $D$, then—to some degree—it is under the control of the decision maker. Consequently, the decision maker can not observe $x$ prior to choosing an act for $D$. For example, in the omelet story, we can not observe any of the responsive variables $o$, $g$, or $s$ before choosing an act.[8]

As an example of an unresponsive variable, suppose we include $S$ (the state of the world) as a variable in $U$. (E.g., in the omelet story, we can take $U$ to be $\{S, o, g, s\}$.) By Savage's definition of $S$, it must be unresponsive to $D$. Note that including $S$ in $U$ creates no new states of the world.

As we have discussed, the notions of unresponsiveness and limited unresponsiveness are closely related to concepts in counterfactual reasoning. When we determine whether or not a chance variable $x$ is unresponsive to decisions $D$, we essentially answer the query "Will the outcome of $x$ be the same no matter how we choose $D$?" Furthermore, when we determine whether or not $x$ is unresponsive to $D$ in states limited by $Y$, we answer the query "If $Y$ will not change as a result of our choice for $D$, will the outcome of $x$ be the same?" One of the fundamental assumptions of our work presented here is that these counterfactual queries are easily answered. In our experience, we have found that decision makers are indeed comfortable answering such restricted counterfactual queries.

---

7. Technically, we should say that $\{o\}$ is responsive to $D$. For simplicity, however, we usually drop set notation for singletons.
8. To be more precise, the variable $o$ represents the number of eggs in the omelet *after* we choose an act for $D$. This variable should not be confused with another variable–say $o'$—corresponding to the number of eggs in the omelet *before* we choose $D$. Whereas $o$ is responsive to $D$ and cannot be observed before choosing an act, $o'$ is unresponsive to $D$ and can be observed before choosing $D$.

The concepts of responsiveness and probabilistic independence are related, as illustrated by the following theorem.

**Theorem 1** *If a set of chance variables $X$ is unresponsive to a set of decision variables $D$, then $X$ is probabilistically independent of $D$.*

**Proof:** By definition of unresponsiveness, $X$ assumes the same instance for all acts in any possible state of the world. Consequently, we can learn about $X$ by observing $S$, but not by observing $D$. □

Nonetheless, the two concepts are not identical. In particular, the converse of Theorem 1 does not hold. For example, let us consider the simple decision of whether to bet heads or tails on the outcome of a coin flip. Assume that the coin is fair (i.e., the probabilities of heads and tails are both 1/2) and that the person who flips the coin does not know our bet. Here, the possible outcomes of the coin toss correspond to the possible states of the world. Further, let decision variable $b$ denote our bet, and chance variable $w$ describe the possible consequences that we win or not. In this situation, $w$ is responsive to $b$, because for both possible states of the world, $w$ will be different for the different bets. Nonetheless, the probability of $w$ is 1/2, whether we bet heads or tails. That is, $w$ and $b$ are probabilistically independent.

Limited unresponsiveness and conditional independence are less closely related than are their unqualified counterparts. Namely, limited unresponsiveness does not imply conditional independence. For example, in the medical-treatment story, $c$ (*cured?*) is unresponsive to $r$ (*recommendation*) in states limited by $t$ (*taken?*), but it is reasonable for us to believe that $c$ and $r$ are not independent given $t$, perhaps because there is some factor that—partially or completely—determines how a person reacts to both recommendations and treatment.

We can derive several interesting properties of limited unresponsiveness from its definition.

1. $X \not\hookleftarrow_Y D \iff \forall x \in X, x \not\hookleftarrow_Y D$

2. $X \not\hookleftarrow_W D \iff X \cup W \not\hookleftarrow_W D$

3. $X \not\hookleftarrow_D D$

4. $X \not\hookleftarrow_Y D \implies X \not\hookleftarrow_{Y \cup Z} D$

5. $X \not\hookleftarrow_{Y \cup Z} D$ and $Y \not\hookleftarrow_Z D \implies X \not\hookleftarrow_Z D$

6. $X \hookleftarrow_Z D$ and $W \not\hookleftarrow_Z D \implies X \hookleftarrow_{W \cup Z} D$

where $D$ is the set of decision variables in the domain, $X$ and $W$ are arbitrary sets of chance variables in $U$, and $Y$ and $Z$ are arbitrary sets of variables in $U \cup D$.

The proofs of these properties are straightforward. For example, consider property 5. Given $X \not\leadsto_{Y \cup Z} D$, we have

$$\forall\, \mathbf{S} \in S,\ \mathbf{D_1} \in D,\ \mathbf{D_2} \in D:\quad Y[\mathbf{S}, \mathbf{D_1}] = Y[\mathbf{S}, \mathbf{D_2}] \text{ and } Z[\mathbf{S}, \mathbf{D_1}] = Z[\mathbf{S}, \mathbf{D_2}]$$

$$\implies X[\mathbf{S}, \mathbf{D_1}] = X[\mathbf{S}, \mathbf{D_2}]$$

Given $Y \not\leadsto_Z D$, we have

$$\forall\, \mathbf{S} \in S,\ \mathbf{D_1} \in D,\ \mathbf{D_2} \in D:\quad Z[\mathbf{S}, \mathbf{D_1}] = Z[\mathbf{S}, \mathbf{D_2}] \implies Y[\mathbf{S}, \mathbf{D_1}] = Y[\mathbf{S}, \mathbf{D_2}]$$

Consequently, we obtain

$$\forall\, \mathbf{S} \in S,\ \mathbf{D_1} \in D,\ \mathbf{D_2} \in D:\quad Z[\mathbf{S}, \mathbf{D_1}] = Z[\mathbf{S}, \mathbf{D_2}] \implies X[\mathbf{S}, \mathbf{D_1}] = X[\mathbf{S}, \mathbf{D_2}]$$

That is, $X \not\leadsto_Z D$.

Other properties follow from these. For example, it is true trivially that $\emptyset \not\leadsto_Y D$. Consequently, by Property 2, we know that $Y \not\leadsto_Y D$. As another example, a special case of Property 4 is that whenever $X$ is unresponsive to $D$, then $X$ will be unresponsive to $D$ in states limited by any $Z$. Also, Properties 4 and 5 imply that limited unresponsiveness is transitive: $X \not\leadsto_Y D$ and $Y \not\leadsto_Z D$ imply $X \not\leadsto_Z D$.

In closing this section, we note that the definition of limited unresponsiveness can be generalized in several ways. In one generalization, we can define what it means for $X \subseteq U$ to be unresponsive to $D$ in states of the world limited by $\mathcal{Y}$, a set of instances of $Y$. Namely, we say that $X$ is unresponsive to $D$ in states limited by $\mathcal{Y}$ if, for all possible states of the world $\mathbf{S}$, and for any two acts $\mathbf{D_1}$ and $\mathbf{D_2}$, $Y[\mathbf{S}, \mathbf{D_1}] = Y[\mathbf{S}, \mathbf{D_2}] \in \mathcal{Y}$ implies $X[\mathbf{S}, \mathbf{D_1}] = X[\mathbf{S}, \mathbf{D_2}]$.

In a second generalization, we can define what it means for a set of chance variables to be unresponsive to a *subset* of all of the decisions. In particular, given a domain described by $U$ and $D$, we say that $X \subseteq U$ is unresponsive to $D' \subseteq D$ in states limited by $Y$ if $X \not\leadsto_{Y \cup (D \setminus D')} D$.

## 3. Definition of Cause

Given the notion of limited unresponsiveness, we can formalize our definition of cause.

**Definition 2 (Causes with Respect to Decisions)** *Given a decision problem described by $U$ and $D$, and a variable $x \in U$, the variables $C \subseteq D \cup U \setminus \{x\}$ are said to be causes for $x$ with respect to $D$ if $C$ is a minimal set of variables such that $x \not\leadsto_C D$.*

In our framework, decision variables can not be caused, because they are under the control of the decision maker. Consequently, we define causes for chance variables only. Also, as we have discussed, our definition is an extension of existing intervention-based definitions of cause (e.g., Rubin [1978]) in that we allow causes to include chance variables. In addition, our definition of cause departs from traditional usage of the term in that cause-effect assertions may vary with the set of decisions available. We discuss the advantages of this departure shortly.

As an example of our definition, consider the decision to continue or quit smoking, described by the decision variable $s$ (*smoke*) and the chance variable $l$ (*lung cancer?*). If we believe that $s$ and $l$ are probabilistically dependent, then, by Theorem 1, it must be that $l \hookleftarrow s$. Furthermore, by Property 3, we know that $l \not\hookleftarrow_s s$. Consequently, by Definition 2, we have that $s$ is a cause of $l$ with respect to $s$.

As another example, consider the medical-treatment story. We have that $c$ (*cured?*) is responsive to $r$ (*recommendation*), because (among other reasons) in the first row in Table 3, the patient is cured if and only if we recommend the treatment. Furthermore, as we discussed in the previous section, $c$ is unresponsive to $r$ in states limited by $t$ (*taken?*). Consequently, we have that $t$ is a cause of $c$ with respect to $r$.

The advantage of defining cause relative to decisions is made clear by our breast-cancer example given in the introduction. Let $g$ and $c$ denote the chance variables *gender?* and *breast cancer?*, respectively. Now, imagine two decisions available to alter gender: $o$, a decision to have a sex-change operation at birth, and $d$, a decision to change chromosomes at conception by microsurgery. It is possible for someone to believe that $c \not\hookleftarrow o$ and yet $c \hookleftarrow d$ and $c \not\hookleftarrow_g d$. That is, it is possible for someone to believe that gender is a cause of breast cancer with respect to the chromosome change but not with respect to the sex-change operation. In this situation, it does not make sense to make the unqualified statement "gender is a cause of breast cancer." In general, our decision-based definition provides added clarity.

Several consequences of Definition 2 are worth mentioning. First, although cause is irreflexive by definition, it is not always asymmetric. For example, in our story about the coin toss, consider another variable $m$ that represents whether or not the outcome of the coin toss matches our bet $b$. In the story as we have told it, $m$ is a deterministic function of $w$ (*win?*), and vice versa. Consequently, we have $w \not\hookleftarrow_m b$ and $m \not\hookleftarrow_w b$; and so $m$ is a cause of $w$ and $w$ is cause of $m$ with respect to $b$. Note that any hint of uncertainty destroys this symmetry. For example, if there is a possibility that the person tossing the coin will cheat (so that we may lose even if we match), then we can conclude that $m$ is a cause of $w$, but not vice versa. This symmetry would also be destroyed if we had a decision controlling $w$ to which $m$ is unresponsive.

Second, cause is transitive for single variables. In particular, if $x$ is a cause for $y$ and $y$ is a cause for $z$ with respect to $D$, then $z \hookleftarrow_D$ and (by the transitivity of unresponsiveness) $z \not\hookleftarrow_x D$. Consequently, $x$ is a cause for $z$ with respect to $D$. Note that transitivity does not necessarily hold for causes containing sets of variables, because the minimality condition in Definition 2 may not be satisfied.

Third, $C = \emptyset$ is a set of causes for $x$ with respect to $D$ if and only if $x$ is unresponsive to $D$.

Fourth, we have the following theorem, which follows from Definition 2 and several of the properties of limited unresponsiveness given in Section 2.

**Theorem 2** *Given any $x \in U$, if $C$ is a set of causes for $x$ with respect to $D$, and $w \in C \cap U$, then $w$ must be responsive to $D$.*

**Proof:** For any chance variable $w \in C$, let $C' = C \setminus \{w\}$. By the minimality condition in our definition, we have

$$x \hookleftarrow_{C'} D \tag{1}$$

Suppose that $w \not\leftharpoonup D$. Then, by Property 4, we have

$$w \not\leftharpoonup_{C'} D \tag{2}$$

Applying Equations 1 and 2 to Property 6, we have that $x \leftharpoonup_C D$, which contradicts that $C$ is a set of causes for $x$ with respect to $D$. □

To illustrate the use of this theorem, let us extend the medical-treatment example by imagining that there is some gene that affects how a person reacts to both our recommendation and to therapy. In this situation, it is reasonable for us to assert that the variable $g$ (*genotype?*) is unresponsive to $r$. Thus, by Theorem 2, $g$ can not be among the causes for any variable.

This consequence of our definition may seem unappealing. Intuitively, we would like to be able to say that (in some sense) $g$ is a cause of $c$. Indeed, our definition does not preclude the ability to make such assertions. Namely, there is no reason to require that the decisions $D$ be implementable in practice or at all. If we want to think about whether or not the patient's genotype is a cause for his cure, then we can imagine an action that can alter one's genetic makeup—for example, retroviral therapy ($v$). In this case, it is reasonable to conclude that $\{r, g\}$ is a cause for $t$ with respect to the decisions $\{r, v\}$. Nonetheless, as we have discussed, we must be clear about the action(s) that alter genotype to make this statement of cause precise.

Finally, we can generalize our definition of what it means for a set of *variables* to cause $x$ to a definition of what it means for a set of *instances* to cause $x$. Namely, we say that $\mathcal{C}$, a set of instances of $C$, is a cause for $x \notin C$ with respect to $D$ if $C$ is a minimal set of variables such that $x$ is unresponsive to $D$ in states limited by $\mathcal{C}$. That is, $\mathcal{C}$ is a cause for $x$ with respect to $D$ if we replace our definition of cause with the weaker requirement that $x$ be unresponsive to $D$ in states limited by $\mathcal{C}$.

## 4. Influence Diagrams

In this and the following three sections, we examine the graphical representation of cause within our framework. This study is useful in its own right, and also will help to relate our framework with Pearl's structural equation model. We begin, in this section, with a review of the influence-diagram representation.

An *influence diagram* is (1) a acyclic directed graph $G$ containing decision and chance nodes corresponding to decision and chance variables, and information and relevance arcs, representing what is known at the time of a decision and probabilistic dependence, respectively, (2) a set of probability distributions associated with each chance node, and optionally (3) a utility node and a corresponding set of utilities (Howard and Matheson, 1981).

An *information arc* is one that points to a decision node. An *information arc* from chance or decision node $a$ to decision node $d$ indicates that variable $a$ will be known when decision $d$ is made. (We shall use the same notation to refer to a variable and its corresponding node in the diagram.) A *relevance arc* is one that points to a chance node. The *absence* of a possible relevance arc represents conditional independence. To identify relevance arcs, we start with an ordering of the variables in $U = (x_1, \ldots, x_n)$. Then, for each variable $x_i$ in order, we ask the decision maker to identify a set $Pa^G(x_i) \subseteq \{x_1, \ldots, x_{i-1}, D\}$ that renders

$x_i$ and $\{x_1, \ldots, x_{i-1}, D\}$ conditionally independent. That is,

$$p(x_i|x_1, \ldots, x_{i-1}, D, \xi) = p(x_i|Pa^G(x_i), \xi) \tag{3}$$

where $p(X|Y, \xi)$ denotes the probability distribution of $X$ given $Y$ for a decision maker with background information $\xi$. For every variable $z$ in $Pa^G(x_i)$, we place a relevance arc from $z$ to $x_i$ in graph $G$ of the influence diagram. That is, the nodes $Pa^G(x_i)$ are the *parents* of $x_i$ in $G$.

Associated with each chance node $x_i$ in an influence diagram are the probability distributions $p(x_i|Pa^G(x_i), \xi)$. From the chain rule of probability, we know that

$$p(x_1, \ldots, x_n|D, \xi) = \prod_{i=1}^{n} p(x_i|x_1, \ldots, x_{i-1}, D, \xi) \tag{4}$$

Combining Equations 3 and 4, we see that any influence diagram for $U \cup D$ uniquely determines a joint probability distribution for $U$ given $D$. That is,

$$p(x_1, \ldots, x_n|D, \xi) = \prod_{i=1}^{n} p(x_i|Pa^G(x_i), \xi) \tag{5}$$

Influence diagrams may also contain special chance nodes. A *deterministic node* corresponds to variable that is a deterministic function of its parents. A *utility node* encodes preferences of the decision maker. Finally, an influence diagram is unambiguous when its decision nodes are totally ordered—that is, when there is a directed path in the influence diagram that traverses all decisions. This total order corresponds to the order in which decisions are made.

In this paper, we concern ourselves neither with the ordering of decision nodes nor the observation of chance variables before making decisions. Therefore, we are not concerned with information arcs. Likewise, although our new concepts apply to models that include utility nodes, we can illustrate these concepts with models containing only chance, deterministic, and decision variables.

Figure 1a contains an influence diagram for the omelet story. As is illustrated in the figure, we use ovals, double ovals, and squares to represent chance, deterministic, and decision nodes, respectively. Among the possible relevance arcs in the influence diagram, several are missing. For example, there is no arc from $D$ to $S$, representing the independence of $D$ and $S$ (which follows from the assertion that $S$ is unresponsive to $D$). Figures 1b and 1c contain influence diagrams for the medical-treatment example. The chance variable $g$ (*genotype?*) is explicitly modeled in Figure 1c.

The ordinary influence diagram was designed to be a representation of conditional independence. Furthermore, as we have discussed, the concepts of conditional independence and limited unresponsiveness are only loosely related. Consequently, the influence diagram is an inadequate representation of causal dependence, at least by our definition of cause.

In particular, an influence diagram may contain an arc from node $x$ to node $y$, even though $x$ is not among a set of causes for $y$. For example, the influence diagram of Figure 1b has an arcs from $r$ and $t$ to $c$ due to the dependencies in the domain. Nonetheless, we have established that the singleton $\{t\}$ is a cause for $c$ with respect to $r$.

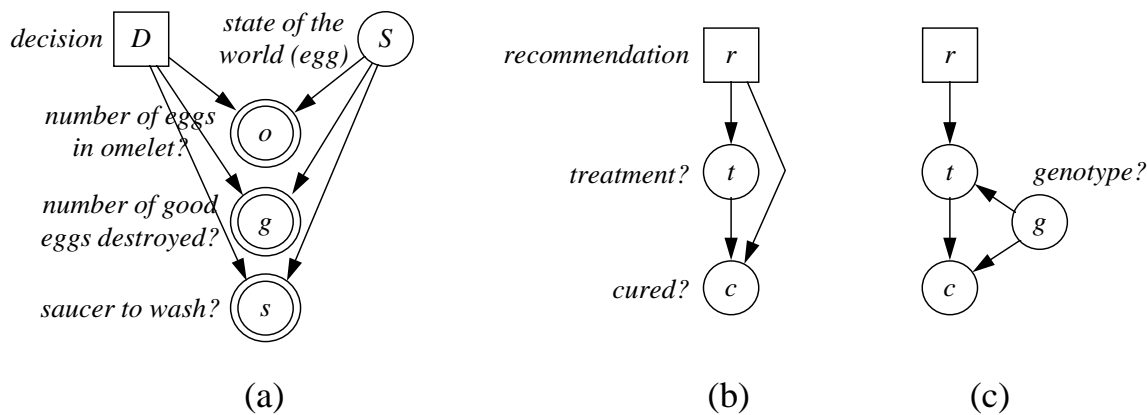

Figure 1: Influence diagrams for (a) the omelet story, and (b,c) the medical-treatment example.

Furthermore, an influence diagram may contain no arc from $x$ to $y$, even though $x$ is a cause of $y$. For example, consider the coin example, illustrated by the influence diagram in Figure 2a. If we believe that the coin is fair, and if we do not bother to model the variable $c$ explicitly (as shown in Figure 2b), then we need not place an arc from $d$ to $w$, because the probability of winning will be $1/2$, regardless of our choice $d$. Nonetheless, $b$ is a cause for $w$ with respect to $b$, by our definition.

Despite these limitations, the influence diagram *is* adequate for purposes of making decisions under uncertainty. In the introduction, we argued that causal information is needed for predicting the effects of actions. Thus, the question arises: "Why do we need anything more than the influence diagram as a representation of the effects of actions?" We give an answer to this question in Section 9, where we discuss counterfactual reasoning. There, we show that the ordinary influence diagram is inadequate for purposes of counterfactual reasoning unless it is in canonical form—a form that accurately reflects cause.

## 5. Direct and Atomic Interventions

In order to define canonical form, we need the concept of a mapping variable. Likewise, in order to define a mapping variable, we need the concept of atomic intervention. We also need the concept of atomic intervention to explicate Pearl's structural-equation model. In this section, we define atomic intervention along with a more general concept called direct intervention.

Roughly speaking, we say that a set of decisions $I$ is a direct intervention on a set of chance variables $X$ if the effects of $I$ on all chance variables are mediated only through the effects of $I$ on $X$. SGS, who take cause to be a primitive, provide a formal definition of direct intervention (which they call a direct manipulation) that is consistent with our notion. We find it simpler to define direct intervention in terms of limited unresponsiveness.

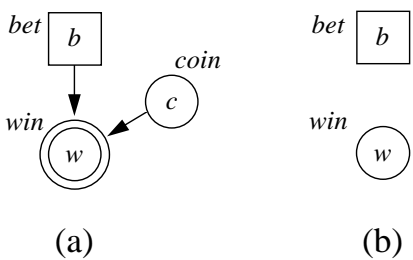

Figure 2: Influence diagrams for betting on a coin flip.

**Definition 3 (Direct Intervention)** *Given a domain described by $U$ and $D$, a set of decisions $I \subseteq D$ is said to be a* direct intervention *on $X \subseteq U$ with respect to $D$ if (1) for all $x \in X$, $x \hookleftarrow I$, and (2) for all $y \in U$, $y \not\hookleftarrow_X I$.*

For example, in the medical-treatment story, $r$ is a direct intervention on $t$, because $t \hookleftarrow r$ and $c \not\hookleftarrow_t r$. As another example, suppose the physician has an additional decision $p$ of whether or not to pay the patient to take the treatment. It is reasonable to expect that $t \hookleftarrow p$. Furthermore, if the amount of payment is small, it is reasonable that $c \not\hookleftarrow_t p$. Consequently, $p$ qualifies as a direct intervention on $t$. Nonetheless, if the amount of payment is sufficiently large, the patient may use that money to improve his health care. Thus, $c \hookleftarrow_t p$; and $p$ does not satisfy the condition 2 for a direct intervention on $t$.

Given the notion of direct intervention, we can define atomic intervention.

**Definition 4 (Atomic Intervention)** *Given a domain described by $U$ and $D$, a decision $\hat{x} \in D$ is said to be an* atomic intervention *on $x \in U$ with respect to $D$ if (1) $\{\hat{x}\}$ is a direct intervention on $\{x\}$ with respect to $D$, and (2) $\hat{x}$ has precisely the instances (a)* **idle**, *which corresponds to the instance of doing nothing to $x$, and (b)* **set**$(\mathbf{x})$ *for every instance $\mathbf{x}$ of $x$, where $x = \mathbf{x}$ whenever $\hat{x} =$* **set**$(\mathbf{x})$.

As we discussed in the introduction, Pearl takes the concept of atomic intervention to be primitive. Whether or not a decision is a direct (or atomic) intervention, however, depends on the underlying causal relationships in the domain. In the medical-treatment story, suppose the physician has a decision $k$ of whether or not to administer the treatment (a drug) without the patient's knowledge. If we believe that the treatment is truly effective and has no placebo effect, then we can assert that $k$ is a direct intervention on $t$. If, however, we believe that the treatment has only a placebo effect, then $k$ will not be a direct intervention on $t$, because $k$ will also directly affect $c$. Thus, the notions of direct and atomic intervention require definitions, lest the meaning of cause would be hidden in these primitives.

We note that, when there are bi-directional causal relationships among variables in $U$, it is not always possible for every chance variable to have its own atomic intervention. For example, consider an adiabatic system consisting of a cylindrical chamber with a moveable top, in which we model the variables *pressure?* ($p$) and *volume?* ($v$).[9] If we allow the top of the chamber to move freely, then placing various weights on the top of the chamber constitutes an atomic intervention on $p$; and we have that $p$ is a cause of $v$ with respect to $\hat{p}$. In contrast, fixing the top of the chamber at particular locations constitutes an atomic intervention on $v$; and we have that $v$ is a cause of $p$ with respect to $\hat{v}$. By the laws of physics, however, both decisions $\hat{p}$ and $\hat{v}$ can not be available simultaneously.

| instance of $t(r)$ | $r$ =take | $r$ =don't take |
|---|---|---|
| 1: complier | $t$ =yes | $t$ =no |
| 2: defier | $t$ =no | $t$ =yes |
| 3: always taker | $t$ =yes | $t$ =yes |
| 4: never taker | $t$ =no | $t$ =no |

Table 4: The mapping variable $t(r)$.

## 6. Mapping Variables

To understand the concept of a mapping variable, let us reexamine Savage's basic formulation of a decision problem. Recall that the chance variables $U$ are a deterministic function of the decision variables $D$ and the state of the world $S$. In effect, each possible state of the world defines a mapping from the decisions $D$ to the chance variables $U$. Thus, $S$ represents all possible mappings from $D$ to $U$. We can characterize $S$ as a mapping variable for $U$ as a function of $D$, and use the suggestive notation $U(D)$ to denote this mapping variable.

In general, given a domain described by $U$, $D$, and $S$, a set of decision variables $Y \subseteq D$, and a set of chance variables $X \subseteq U$, the mapping variable $X(Y)$ is a variable that represents the possible mappings from $Y$ to $X$.

As an example, consider the medical-treatment story. The mapping variable $t(r)$ represents the possible mappings from the decision variable $r$ (*recommendation*) to the chance variable $t$ (*taken?*). In this example, the instances of $t(r)$, shown in Table 4, have a natural interpretation. In particular, the instance where the patient accepts treatment if and only if we recommend it represents a patient who complies with our recommendation; the instance where the patient accepts treatment if and only if we recommend against it represents a patient who defies our recommendation; and so on.

The notion of a mapping variable is discussed in Heckerman and Shachter (1994), and in Balke and Pearl (1994) under the name "response function." A related counterfactual variable is described by Neyman (1923), Rubin (1978), and Howard (1990). They discuss what we would denote $X(Y = \mathbf{Y})$: the variable $X$ *if* we choose instance $\mathbf{Y}$ for $Y$.

An important property concerning mapping variables is that, given variables $X, Y$, and $X(Y)$, we can always write $X$ as a deterministic function of $Y$ and $X(Y)$. For example, $t$ is a deterministic function of $r$ and $t(r)$; and $U$ is a deterministic function of $D$ and $U(D) \equiv S$.

---

9. This example is not appropriate technically, as it uses continuous variables. Nonetheless, this example illustrates our point.

In the discussions that follow, it is useful to extend the definition of a mapping variable to include chance variables as arguments. For example, in the medical-treatment story, it seems reasonable to define the mapping variable $c(t)$ with instances **helped**, **hurt**, **always cured**, and **never cured**. Together, the mapping variables $t(r)$ and $c(t)$ describe the possible states of the world $U(D) \equiv S$. (E.g., $t(r) =$**complier** and $c(t) =$**helped** corresponds to state 1 in Table 3.) As we shall see, this decomposition of $U(D)$ facilitates the graphical representation of causal relationships.

Unfortunately, defining mapping variables with chance-variable arguments is not always possible. In the medical-treatment domain, when the patient is an always taker (states 10 and 11 in Table 3), $t=$**yes** regardless of $r$. Consequently, we can not tell whether $c(t)$ is **helped** or **always cured**—that is, $c(t)$ is not uniquely identified. Because Savage's decision-theoretic framework requires that the state of the world and the act uniquely determine the instance of $c(t)$ (a consequence), the instance of $c(t)$ is not well defined. Nonetheless, $c(t)$ is well defined whenever $D$ includes an atomic intervention on $t$ ($\hat{t}$), guaranteeing that $t$ will take on all instances (as $\hat{t}$ varies) in every state of the world.

In general, we have the following definition of mapping variable.

**Definition 5 (Mapping Variable)** *Given a domain described by $U$ and $D$, chance variables $X$, and variables $Y$ such that, for every $y \in Y \cap U$, there exists an atomic intervention $\hat{y} \in D$,[10] the mapping variable $X(Y)$ is the chance variable that represents all possible mappings from $Y$ to $X$.*

There are several important points to be made about mapping variables as we have now defined them. First, as in the more specific case, $X$ is always a deterministic function of $Y$ and $X(Y)$.

Second, additional probability assessments typically are required when introducing a mapping variable into a probabilistic model. For example, two independent assessments are needed to quantify the relationship between $r$ and $t$ in the medical-treatment story; whereas three independent assessments are required for the node $t(r)$. In general, many additional assessments are required. If $X$ has $c$ instances and $Y$ has $a$ instances, then $X(Y)$ has as many as $c^a$ instances. In real-world domains, however, reasonable assertions of independence decrease the number of required assessments. In some cases, no additional assessments are necessary (see, e.g., Heckerman et al., 1994).

Third, we have the following theorem, which follows immediately from the definitions of limited unresponsiveness and mapping variable. In this and subsequent theorems that mention mapping variables, we assume that atomic interventions required for the proper definition of the mapping variables are included in $D$.

**Theorem 3 (Mapping Variable)** *Given a decision problem described by $U$ and $D$, variables $X \subseteq U$, and $Y \subseteq U \cup D$, $X \not\leftarrow_Y D$ if and only if $X(Y) \not\leftarrow D$.*

For example, in the medical treatment domain that includes the atomic intervention $\hat{t}$, we have $c \not\leftarrow_t \{r, \hat{t}\}$ and $c(t) \not\leftarrow \{r, \hat{t}\}$. Roughly speaking, Theorem 3 says that $X$ is unresponsive to $D$ in states limited by $Y$ if and only if the way $X$ depends on $Y$ does not depend on $D$. This equivalence provides us with an alternative set of conditions for cause.

---

10. Recall from Section 5 that it is not always possible to have atomic interventions for every $y \in Y$.

**Corollary 4 (Causes with Respect to Decisions)** *Given a decision problem described by $U$ and $D$, and a chance variable $x \in U$, the variables $C \subseteq D \cup U \setminus \{x\}$ are causes for $x$ with respect to $D$ if only if $C$ is a minimal set of variables such that $x(C) \not\hookleftarrow D$.*

When $C$ are causes for $x$ with respect to $D$, we call $x(C)$ a *causal mapping variable with respect to $D$*. Thus, we have the following consequence of Theorem 3.

**Corollary 5 (Causal Mapping Variable)** *If $x(C)$ is a causal mapping variable for $x$ with respect to $D$, then $x(C)$ is unresponsive to $D$.*

## 7. Canonical Form Influence Diagrams

We can now define what it means for an influence diagram to be in canonical form.

**Definition 6 (Canonical Form)** *An influence diagram for a decision problem described by $U$ and $D$ is said to be in* canonical form *if (1) all chance nodes that are responsive to $D$ are descendants of one or more decision nodes and (2) all chance nodes that are descendants of one or more decision nodes are deterministic nodes.*

An immediate consequence of this definition is that any chance node that is not a descendant of decision node must be unresponsive to $D$.

We can construct an influence diagram in canonical form for a given problem by including in the influence diagram a causal mapping variable for every variable that is responsive to the decisions. In doing so, we can make every responsive variable a deterministic function of its mapping variable and the corresponding set of causes. For example, consider the medical-treatment story as depicted in the influence diagram of Figure 3a. The variables $t$ and $c$ are responsive to $r$, but their corresponding nodes are not deterministic. Consequently, this influence diagram is not in canonical form. To construct a canonical form influence diagram, we introduce the mapping variables $t(r)$ and $c(r)$, as shown in Figure 3b. The responsive variables are now deterministic; and the mapping variables are unresponsive to the decision. This example illustrates an important point: Mapping variables may be probabilistically dependent. We return to this issue in Section 8.

In general, we can construct an influence diagram in canonical form for any decision problem characterized by $U$ and $D$ as follows.

**Algorithm 1 (Canonical Form)**

1. *Add a node to the diagram corresponding to each variable in $U \cup D$*

2. *Order the variables $x_1, \ldots, x_n$ in $U$ so that the variables unresponsive to $D$ come first*

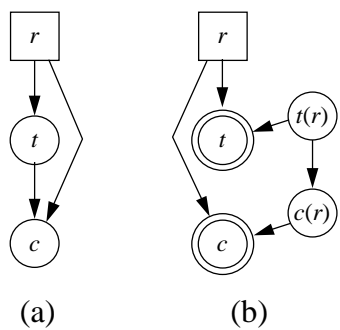

Figure 3: (a) An influence diagram for the medical-treatment story. (b) A corresponding influence diagram in canonical form.

3. For each variable $x_i \in U$ that is responsive to $D$,

   (a) Add a causal-mapping-variable chance node $x_i(C_i)$ to the diagram, where $C_i \subseteq D \cup \{x_1, \ldots, x_{i-1}\}$

   (b) Make $x_i$ a deterministic node with parents $C_i$ and $x_i(C_i)$

4. Assess independencies among the variables that are unresponsive to $D$[11]

This algorithm is well defined, because it is always possible to find a set $C_i$ satisfying the condition in step 3a. In particular, $x_i \not\hookrightarrow_D D$ by Property 3. Consequently, even when $D$ contains no atomic intervention, we can always create a causal mapping variable for every responsive variable in $U$.

Also, the structure of any influence diagram constructed using Algorithm 1 will be valid. Namely, by Corollary 5, all causal mapping variables added in step 3 are unresponsive to $D$. Thus, suppose we identify the relevance arcs and deterministic nodes according to Equation 3 by using a variable ordering where the nodes in $D$ are followed by the unresponsive nodes (including the causal mapping variables), which are in turn followed by the responsive nodes in the order specified at step 2. Then, (1) we would add no arcs from $D$ to the unresponsive nodes by Theorem 1 (and the algorithm adds none); (2) we would add arcs among the unresponsive nodes as described in step 4; and (3) for every responsive variable $x_i$, we would make $x_i$ a deterministic node (as described in step 3b) by definition of a mapping variable.

In addition, the structure that results from Algorithm 1 will be in canonical form. In particular, because there are no arcs from $D$ to the unresponsive nodes, only responsive variables can be descendants of $D$. Also, by Theorem 2, we know that every responsive node is a descendant of $D$, and (by construction) a deterministic node.

---

11. Because mapping variables are random variables, the assessment of dependencies among the unresponsive variables is, in principle, no different than that for assessing dependencies among ordinary random variables. Nonetheless, the counterfactual nature of the variables can be confusing. Howard (1990) describes a method of probability assessment that addresses this concern.

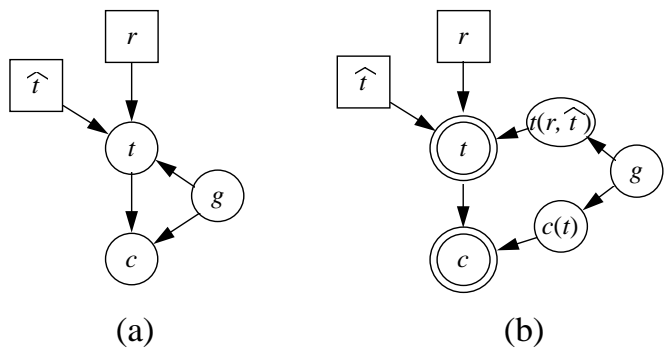

Figure 4: (a) Another influence diagram for the medical-treatment story. (b) A corresponding influence diagram in canonical form.

Furthermore, by construction, every responsive variable $x_i \in U$ has one set of causes explicitly encoded in the diagram ($C_i$).

To illustrate the algorithm, consider the medical-treatment story as depicted by the influence diagram in Figure 4a, where the variable $g$ (*genotype?*) is represented explicitly, and where $c \not\hookrightarrow_t \{r,\hat{t}\}$ and $g \not\hookrightarrow \{r,\hat{t}\}$. To construct an influence diagram in canonical form for this problem, we first add the variables $\{r,\hat{t},g,t,c\}$ to the diagram and choose the ordering $(g,t,c)$. Both $t$ and $c$ are responsive to $D = \{r,\hat{t}\}$, and have causes $\{r,\hat{t}\}$ and $t$, respectively. Consequently, we add causal mapping variables $t(r,\hat{t})$ and $c(t)$ to the new diagram, and make $t$ a deterministic function of $r$, $\hat{t}$, and $t(r,\hat{t})$ and $c$ a deterministic function of $t$ and $c(t)$. Finally, we assess the dependencies among the unresponsive variables $\{g,t(r,\hat{t}),c(t)\}$, adding arcs from $g$ to $t(r,\hat{t})$ and $c(t)$ under the assumption that the causal mapping variables are conditionally independent given $g$. The resulting canonical form influence diagram is shown in Figure 4b.

Canonical form is a generalization of Howard Canonical Form, which was developed by Howard (1990) to facilitate the computation of value of information.[12] Before making important decisions, decision analysts investigate how useful it is to gather additional information. This investigation is typically done by computing the extra value the decision maker would obtain by observing earlier one or more chance variables in the domain. If the decision maker does not expect to observe chance variable $x$ prior to making decision $d$, the *value of information about $x$* is the extra value he would obtain if he were able to observe chance variable $x$ just before making decision $d$. The value of information is never negative, and it serves as a bound on the value of any experiment: it would never be worthwhile to spend more than the value of information about $x$ to obtain any (possibly imperfect) observation about $x$ just before making decision $d$.

Given an ordinary influence diagram, we can not compute the value of information about variables responsive to $D$, because such variables can not be observed before decisions $D$ are made. In contrast, we can always compute the value of information about mapping variables corresponding to responsive variables in a canonical form influence diagram, because such variables are unresponsive to $D$ by definition. For example, consider the decision to continue or quit smoking described by decision variable $s$ (*smoke*) and chance variables $l$ (*lung cancer?*) and $l(s)$. Although we cannot compute the value of information about $l$ because it is responsive to $D$, we can compute the value of information about $l(s)$.

---

12. Influence diagrams in HCF do not allow mapping variables whose arguments contain chance variables.

At first glance, it may seem pointless to determine the value of information about a variable that cannot be observed (such as $l(s)$). Nonetheless, we can often learn *something* about a mapping variable. For example, imagine a test that measures the susceptibility of someone's lung tissue to lung cancer in the presence of tobacco smoke. Learning the result of such a test may well update our probability distribution over $l(s)$. By computing the value of information of $l(s)$, we obtain an upper bound on the most we would be willing to pay to undergo such a test.

## 8. Pearl's Causal Framework

We can now demonstrate the relationship between Pearl's causal framework and ours. As mentioned, Pearl's framework is similar to that of SGS (see the background notes in SGS for a discussion). Thus, many of the remarks in the section apply to SGS's model for cause as well. A notable exception is that SGS formally define direct intervention.

The following theorem outlines the relationship.

**Theorem 6** *Given chance variables $U$, suppose the set of decision variables $D$ contains a unique atomic intervention $\hat{x}$ for every $x \in U$ and no other decisions. Given graph $G$, a directed acyclic graph with nodes corresponding to the variables in $U$, suppose that, for all $x \in U$, $Pa^G(x) \cup \{\hat{x}\}$ are causes for $x$ with respect to $D$.[13] Then, the relationships among the variables in $U \cup D$ can be expressed by the set of simultaneous equations*

$$x = f_x(Pa^G(x), \hat{x}, x(Pa^G(x), \hat{x}))$$

*for all $x \in U$, where $f_x$ is a deterministic function such that $x = \mathbf{x}$ if $\hat{x} =\mathbf{set}(\mathbf{x})$.*

**Proof:** The theorem follows by applying Algorithm 1 using an ordering over $U$ consistent with the graph $G$. □

Thus, we see that Pearl's structural-equation model is a specialization of canonical form when we identify (1) Pearl's domain variables with our chance variables $U$, (2) Pearl's atomic interventions with our atomic interventions $D$, (3) Pearl's causal graph with our graph $G$, and (4) Pearl's random disturbance $\epsilon_x$ with our causal mapping variable $x(Pa^G(x), \hat{x})$.

This correspondence permits several clarifications of Pearl's framework. First, we have a precise definition of atomic intervention. Unlike Pearl's model, where the concept of atomic intervention is primitive, our framework provides a way to verify that interventions are indeed atomic.

Second, we see what it means for the random disturbances to be exogenous. Namely, these random variables are *unresponsive* to the decisions $D$.

---

13. It is not difficult to show that this condition is consistent with the condition that, for all $x \in U$, $\hat{x}$ is an atomic intervention on $x$.

Third, we have a precise definition of random disturbance in terms of causal mapping variable. Consequently, we have a means for assessing the joint probability distribution of these variables, and—in particular—a means for assessing independencies among these variables. In fact, whereas Pearl requires that random disturbances be marginally independent, our definition imposes no such requirement.

Theorem 6 shows that any structural-equation model can be encoded as an influence diagram in canonical form. The converse is also true—that is, any influence diagram in canonical form can be encoded as a structural-equation model. This result may seem surprising, because in Pearl's model every domain variable must have an atomic intervention, all decision variables must be atomic interventions, and random disturbances must be independent. Given an influence diagram in canonical form, however, we can encode its chance and decision variables in a structural equation model. Specifically, a chance variable $x$ can be encoded as the variable pair $\{x, \hat{x}\}$ where $\hat{x}$ is instantiated to **idle**, and a decision variable $d$ can be encoded as the variable pair $\{d, \hat{d}\}$ where the act **idle** is forbidden. In addition, as noted by Pearl, we can remove dependencies among mapping variables (at least in practice) by introducing hidden common causes.[14]

Nonetheless, because hidden common causes sometimes need to be introduced, Pearl's structural-equation model can be a less efficient representation than canonical form. For example, to represent the relationships in Figure 4b, we would use a structual-equation model with disturbance variables corresponding to $g(\hat{g})$, $t(r, g, \hat{t})$, and $c(t, g, \hat{c})$. Assuming $r, g, t$ and $c$ are binary variables, the disturbance variables have 2, 16, and 16 instances, respectively.[15] Assuming the disturbance variables are independent, the joint probability distribution of these variables contain 31 probabilities. In contrast, both mapping variables in Figure 4b have only four instances. Consequently, the joint probability distribution over the unresponsive variables in the canonical-form representation contain only 13 probabilities.

We note that Balke and Pearl (1994) relax the assumption that mapping variables are independent. Nonetheless, their generalization of the structural-equation model, which they call a *functional model*, is still less efficient than canonical form. The inefficiency comes from the fact that canonical form encodes a joint probability distribution among all unresponsive variables (possibly including both domain and mapping variables), whereas a functional model encodes a joint probability distribution among mapping variables only. For example, the canonical-form influence diagram in Figure 4b encodes the assertion that $t(r, \hat{t})$ and $c(t)$ are independent given $g$. This assertion can not be encoded in the Balke-Pearl representation. When we represent the relationships in Figure 4b using a functional model, we can include the variable $g$, in which case we obtain the 31-probability model described in the previous paragraph. Alternatively, we can exclude variable $g$ from the model, and encode the dependency between the mapping variables $t(r, \hat{t})$ and $c(t, \hat{c})$ with an arc between these two variables. The resulting Balke-Pearl model has 15 probabilities in contrast to the 13 required by canonical form.

---

14. The assumption that the mapping variables are independent has the convenient consequence that the graph $G$ can be interpreted as a Bayesian network in the traditional sense. That is, if variables $X$ and $Y$ are d-separated by $Z$ in $G$, then $X$ and $Y$ are conditionally independent given $Z$ according to the structural-equation model corresponding to $G$. (See Pearl, 1988, for a definition of d-separation.) SGS (p. 54) refer to this association as the *causal Markov condition.*
15. Note that the mapping variable $x(Y, \hat{x})$ has the same number of instances as does the mapping variable $x(Y)$.

## 9. Counterfactual Reasoning

As we have noted, the ordinary influence diagram is adequate for making decisions under uncertainty, but is inadequate for counterfactual reasoning. In this section, we examine this form of reasoning and suggest how it can be facilitated by influence diagrams in canonical form.

Given a domain described by $U$ and $D$ with $X, Y, Z \subseteq U$, counterfactual reasoning addresses questions of the form: If we choose $D = \mathbf{D_1}$ and observe $X = \mathbf{X}$, what is the probability that $Y = \mathbf{Y}$ if we choose $D = \mathbf{D_2}$ and observe $Z = \mathbf{Z}$? For example, in the medical-treatment domain, we may wish to know: If we recommend the treatment and the patient takes the drug and is cured, what is the probability that the patient will be cured if we recommend against the treatment? Such reasoning is often important in the real-world—for example, in legal argument (Ginsberg, 1986; Balke and Pearl, 1994; Goldszmidt and Darwiche, 1994; Heckerman et al., 1994).

We can answer such queries using influence diagrams in canonical form. To illustrate this approach, consider the medical-treatment question in the previous paragraph. To answer this query, we begin with the influence diagram in canonical form shown in Figure 4b. Then, we duplicate all decision variables and all chance variables that are responsive to the decisions, as shown in Figure 5. The original variables represent the act $r =$**take**, $\hat{t}=$**idle** and its consequences. The duplicate variables (denoted with primes) represent the act $r' =$**don't take**, $\hat{t}=$**idle** and its consequences. There is no need to duplicate the unresponsive variables (including the causal mapping variables) because, by definition, they can not be affected by the decisions.[16] Next, we copy the deterministic function associated with each original variable to its primed counterpart. Then, we instantiate the decision and chance variables as described in the query ($r =$**take**, $\hat{t}=$**idle**, $t =$**taken**, $c =$**cured**, $r' =$**don't take**, and $\hat{t}'=$**idle**). Finally, we use a standard Bayesian-network inference method to compute the probability of the variable(s) of interest ($c'$ in our example).

The canonical form influence diagram is a natural representation for counterfactual reasoning for two reasons. One, the deterministic relationships between a responsive chance variable and its parents remains the same for any choice of $D$. Two, the instances assumed by unresponsive variables are unaltered by the decisions. The ordinary influence diagram offers neither of these guarantees.

Our approach, described in Heckerman and Shachter (1994), is similar to that of Balke and Pearl (1994). The main difference between the two approaches is that Balke and Pearl use their functional model as the base representation, making their approach less efficient than ours. Goldszmidt and Darwiche (1994) describe a graphical language for modeling the evolution of real-world systems over time. Although their approach does not explicitly address counterfactual reasoning, it can be adapted to so do, yielding an alternative to our approach.

---

16. In general, we need duplicate only (1) those decision variables that change in the query and (2) those chance variables that are responsive to the decisions that change.

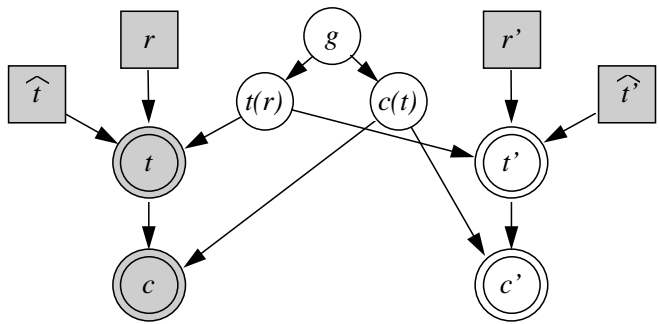

Figure 5: The use of canonical form to compute a counterfactual query. Shaded variables are instantiated.

## 10. Conclusions

We have presented a definition of cause and effect in terms of the decision-theoretic primitives of act, state of the world, and consequence determined by act and state of the world, and have shown how this definition provides a foundation for causal reasoning. Our definition departs from the traditional view of causation in that our causal assertions are made relative to a set of decisions. Consequently, as we have argued, our definition allows for a more precise specification of causal relationships.

In addition, we have shown how our definition provides a basis for the graphical representation of cause. We have described a special class of influence diagrams, those in canonical form, and have shown that it is equally expressive and more efficient than Pearl's structural-equation model. Finally, we have shown how influence diagrams in canonical form, unlike ordinary influence diagrams, can be used for counterfactual reasoning.

## Acknowledgments

We thank Jack Breese, Tom Chavez, Max Chickering, Eric Horvitz, Ron Howard, Christopher Meek, Judea Pearl, Mark Peot, Glenn Shafer, Peter Spirtes, Patrick Suppes, and anonymous reviewers for useful comments.